\documentclass[conference]{IEEEtran}
\IEEEoverridecommandlockouts
\usepackage{cite}
\usepackage{amsmath,amssymb,amsfonts}
\usepackage{algorithmic}
\usepackage{graphicx}
\usepackage{textcomp}
\usepackage{xcolor}
\usepackage{multirow}
\usepackage[utf8]{inputenc}
\usepackage{CJKutf8}

\def\BibTeX{{\rm B\kern-.05em{\sc i\kern-.025em b}\kern-.08em
    T\kern-.1667em\lower.7ex\hbox{E}\kern-.125emX}}
\begin{document}
\begin{CJK}{UTF8}{gbsn}


\title{Multi-level Encoder with Global Topic for Task-oriented Dialogue Summarization}

\author{\IEEEauthorblockN{Zhuoqi He, Peijie Huang\textsuperscript{*}, Yuhong Xu, Youming Peng, Mingzhi Xu, Xinyang Lin}
\IEEEauthorblockA{\textit{College of Mathematics and Informatics, South China Agricultural University,
Guangzhou, China}
\\zhuoqihe497@163.com, pjhuang@scau.edu.cn, xuyuhong@scau.edu.cn,\\
youmingpeng540@163.com, xdxulol@163.com, yizhiyang0111@gmail.com
}\thanks{\textsuperscript{*}Corresponding author.} 
}

\maketitle

\begin{abstract}
Task-oriented dialogue summarization aims to automatically extract key information to generate domain summaries to improve service efficiency and quality. Task-oriented dialogue is inherently logical and surrounds specific topic. How to effectively capture the dialogue topic and the most salient information becomes one of the major challenges of this task. In this paper, we propose a task-oriented dialogue summarization model utilizing a multi-level encoder with global topic (MLEGT). This model discovers temporal and spatial dependencies through a multi-level encoder, while simultaneously obtaining a simple yet effective global topic in the process of graph construction. It not only captures the crucial details but also breaks the long-range limit and discover the intrinsic structure across the dialogue. Then, a global topic guided pointer mechanism is added into the summrizer to capture ignored but important information, which is helpful to improve the accuracy of the summary. A comprehensive study of two public datasets, including a real-world large-scale Chinese police interrogation dataset and a medical reporting service dataset proves the superiority of our method on several strong baselines.
\end{abstract}


\section{Introduction}
In an active industry service system, massive domain-specific task-oriented spoken dialogues conveying important information between human and human/agent are generated in real-time. Automatically summarizing such task-oriented dialogues may aid the archiving, indexing and retrieval of various records of oral communication such as sales interactions, medical diagnosis, complain report to improve the service. Therefore, how to transform a large number of task-oriented dialogues into natural, concise, and core-informed texts such as task-oriented dialogue summaries is gaining significantly increased attention\cite{zou2021,DBLP:journals/corr/abs-2110-12680,sdds2023}. 

\setlength{\textfloatsep}{8pt plus 3pt minus 1pt}
\begin{figure}[h]
\centering
\includegraphics[width=3.6in]{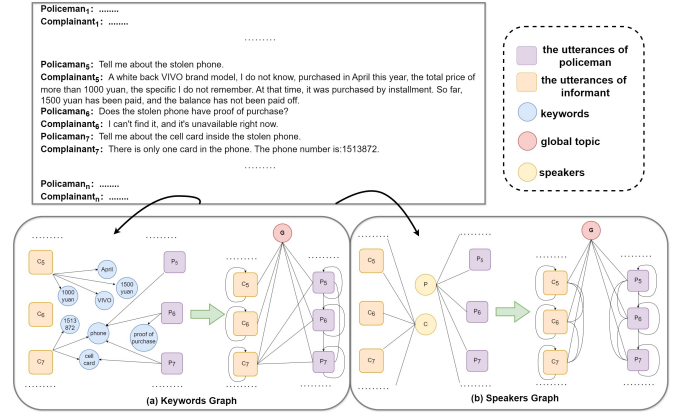}
\caption{The Construction of Graphs.}
\end{figure}

Most of the existing dialogue summary works are mainly focused on open-domain dialogues, like chitchat\cite{DBLP:conf/icassp/LeiYZHZX21,DBLP:conf/emnlp/LeiZYHX21,DBLP:conf/acl/LiPHGYG23,topic2023}, meeting\cite{DBLP:conf/www/ZhaoPFLLY19,DBLP:conf/acl/LiZJR19,DBLP:journals/ipm/SunYLCL24}, which are usually summarized by stringing all dialogue points to maintain an integral conversation flow\cite{DBLP:conf/acl/VazirgiannisTMS18}. Nevertheless, in task-oriented dialogues, dialogue speakers commonly have strong and clear motivations and aim to address issues about a specific task. To this end, recent works focus on developing topic or structure-aware models for task-oriented dialogue summarization. Zou et al.\cite{zou2021} leverage topic information that flows in the dialogue to help generate topic-aware summaries. Liu et al. \cite{DBLP:conf/kdd/LiuWXLY19} introduce a key point sequence as auxiliary label, which helps the model learn the logic of the summary. Zhao et al. \cite{DBLP:journals/corr/abs-2110-12680} 
propose a dialogue state-aware summarization model to combine original dialogue text and structured dialogue state. 
However, these approaches remain at a coarse level in exploring the intrinsic structural features of dialogue. Gao et al.\cite{sdds2023} propose a graph-based dialogue summarization model that uses structure fusion graph to capture the dialogue structure. However, this
method is lacking in topic modeling.

\setlength{\textfloatsep}{8pt plus 2pt minus 1pt}
\begin{figure*}[t]
\centering
\includegraphics[width=5.9in]{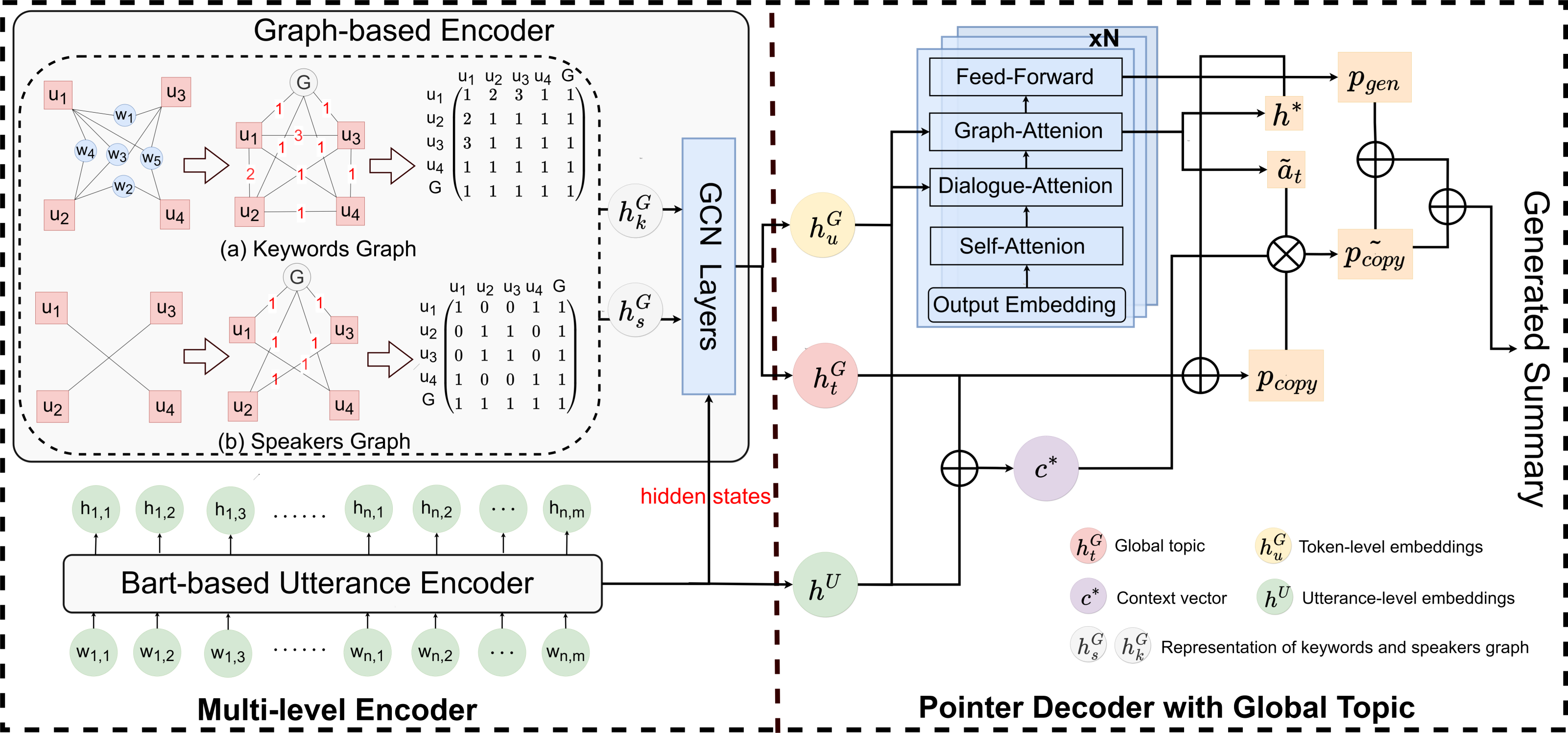}
\caption{Overall Framework of MLEGT.}
\end{figure*}

To address the above issues, we design a graph-based task-oriented summarization model using multi-level encoder with global topic (MLEGT).
A multi-level encoder is proposed to discover the temporal and spatial dependencies across the dialogue.
Simultaneously a simple yet effective global topic is obtained in the process of graph construction, which helps us grasp the flow of information and discover the intrinsic structure across the dialogue. Then, we incorporate a global topic guided pointer mechanism into the summrizer to capture the ignored but important information while generating summary, which reduce the errors for better summarization. Figure 1 shows the construction of graphs how to capture the specific details across the dialogue. The experiments show that MLEGT achieves new state of the art on two public datasets,  including a real-world large-scale Chinese police interrogation dataset and a medical reporting service dataset

\section{Proposed Model}
In this section, we introduce MLEGT model using multi-level encoder (Subsection II.A and II.B) with global topic guided pointer mechanism (in Subsection II.C) for task-oriented summarization. An overview is shown in Figure 2.

\subsection{Utterance Encoder}
We employ the pre-trained BART \cite{bart2020} to encode each utterance independently:
$\setlength\abovedisplayskip{3pt}
\setlength\belowdisplayskip{3pt}
\{h_{i,0},h_{i,1},...,h_{i,L_{u}^{i}}\}=Enc([CLS],w_{i,1},...,w_{i,L_{u}^{i}})
$, where Enc(·) is the encoder module in BART which outputs the vector representation $h_{i,j}$ of $j$-th input word $w_{i,j}$ in $i$-th utterance. 
Then we obtain the token-level Utterance $h^{U}=\{u_{1},...,u_{L_{d}}\}$ in this module.
\subsection{Graph Encoder}

\subsubsection{Graph Construction} 
In this subsection, following Gao et al.\cite{sdds2023}, we construct two types of graphs based on keywords and speakers, which is the Keywords Graph and the Speaker Graph respectively. We propose a new method in graph construction for effectively ensuring the connectivity of the graph. Our method is signiﬁcantly different from Gao et al.\cite{sdds2023} in the following two points: (1) We set the diagonal of the adjacency matrix be 1 so that each utterance can not only discover the relations among their neighbors but also focus on themselves. (2) We add a global node into each graph, which can help us obtain a simple but effective global topic in the process of graph construction. 

\textbf{Keywords Graph.} The construction of Keywords Graph aims to discover the spatial dependencies among different utterances in the dialogue. Firstly, we obtain an utterances-keywords adjacent matrix $A$ based on keywords extracted from utterances by SpaCy.
Each value $A_{i,j}$ in the adjacent matrix corresponds the frequency of the $j$-th keyword occurrences in $i$-th utterance. Let $A$ multiply by its own transpose matrix, and then we obtain the initial Keywords Graph $G_{k}$,
\begin{equation}
\setlength\abovedisplayskip{3pt}
\setlength\belowdisplayskip{3pt}
G_{k}=AA^{T},
\end{equation}

In addition, we set the diagonal of the Keywords Graph to 1, enabling each utterance to also pay attention to itself.

\textbf{Speakers Graph.} The construction of Speaker Graph aims to effectively gather the information of the same speaker and reduce the dispersion of information. Firstly, we obtain an utterances-speakers adjacent matrix $B$ based on different speakers in the dialogue. Each value $B_{i,j}$ in the adjacent matrix corresponds the specific speaker in $i$-th utterance. Here, we use a hot-dot to represent each speaker. Let $B$ multiply by its own transpose matrix, and then we obtain the initial Speakers Graph $G_{s}$,
\begin{equation}
\setlength\abovedisplayskip{3pt}
\setlength\belowdisplayskip{3pt}
G_{s}=BB^{T},
\end{equation}

In addition, we set the diagonal of the Speakers Graph to 1, enabling each utterance to also pay attention to itself.

\textbf{Global Node.} Furthermore, we incorporate a global node into each graph, which is a hyper node \cite{feng2020} connected to all nodes, to aggregate non-local information across the dialogue. Each node builds relationships with its neighboring nodes through edges, while a global node connects to all nodes, serving as a central hub for aggregating global information. It can interact with all nodes and capture important features in the process of information propagation, thereby enhancing the model's overall understanding and modeling capability of the graph structure.

\subsubsection{Graph update}
Graph encoder\cite{feng2021} is used to digest the structural information and get updated node representations. The token-level embeddings $h^{U}$ from the Utterance Encoder is added as the node representations for each graph. Given the constructed $G_{k}=(V_{k},E_{k})$ and $G_{s}=(V_{s},K_{s})$, nodes $V$ correspond to the utterances for each dialogue while edges $E$ correspond to the similarity relations.

\textbf{Utterance-Level embeddings.} To update the representations of all the nodes, we apply a Graph Convolutional Network\cite{gcn2017} to the Keywords Graph and Speaker Graph respectively. 
\begin{equation}
\setlength\abovedisplayskip{3pt}
\setlength\belowdisplayskip{3pt}
h_{k}^{G}=GCN(Add\&Norm(h^{U},\sum_{j \in N_{i}} \frac{1}{|N_{i}|}W_{1}V_{j}),
\end{equation}
\begin{equation}
\setlength\abovedisplayskip{3pt}
\setlength\belowdisplayskip{3pt}
h_{s}^{G}=GCN(Add\&Norm(h^{U},\sum_{j \in N_{i}} \frac{1}{|N_{i}|}W_{2}V_{j}),
\end{equation}
where $Add\&Norm$ denotes residual connection and layer normalization while $N_{i}$ denotes the neighborhood of the $i$-th node. $h_{k}^{G}$ and $h_{s}^{G}$ are the outputs from the Utterance Encoder. 

Since GCN does not share parameters, these two graphs together are then concatenated as $h^{G}_{c}$，
\begin{equation}
\setlength\abovedisplayskip{3pt}
\setlength\belowdisplayskip{3pt}
h^{G}_{c}=concat[h_{k}^{G}[:-1];h_{s}^{G}[:-1]],
\end{equation}

 where $h_{k}^{G}[:-1]$ and $h_{s}^{G}[:-1]$ are the representations of all nodes except the global node for each graph respectively.
 Eventually, it is passed into the cross attention layer, which utilize the dual cross attention mechanism\cite{dualattention2022} across the graphs to further explore the long-range dependencies and uncover the intrinsic structure. And we obtain the utterance-level embeddings,
\begin{equation}
\setlength\abovedisplayskip{3pt}
\setlength\belowdisplayskip{3pt}
h^{G}_{u}=CrossAttn(Add\&Norm(h^{G}_{c}),h^{U},h^{U}),
\end{equation}

\textbf{Global Topic Representation.} 
To better capture the flow of information and uncover the intrinsic structure across the dialogue, we introduced an additional global node that connects to all other nodes. It not only captures the implicit relationships among other nodes, but also forms a representation of each node's contribution to the overall theme. It dynamically updates alongside the other nodes and we regard it as a global topic representation:
\begin{equation}
\setlength\abovedisplayskip{3pt}
\setlength\belowdisplayskip{3pt}
h^{G}_{t}=h_{k}^{G}[-1]+h_{s}^{G}[-1],
\end{equation}
where $h_{k}^{G}[-1]$ and $h_{s}^{G}[-1]$ are the representations of the global node for each graph respectively.

\subsection{Global Topic Guided Pointer Decoder}
\subsubsection{Global Topic Guided Pointer Mechanism}
A pointer mechanism\cite{pointer2017} is added into our model, aiming to capture the ignored but important information during the feature extracting and training.

Firstly, we obtain the last hidden states $h^{*}$ from the decoder. And we project $h^{*}$ and $h^{G}_{t}$ into two new vector spaces separately to get the $P_{copy}$.
\begin{equation}
\setlength\abovedisplayskip{3pt}
\setlength\belowdisplayskip{3pt}
P_{copy}=sigmoid(W_{3}h^{*}+W_{4}h^{G}_{t}),
\end{equation}

Then, we utilize the token-based embeddings $h^{U}$ and the global topic $h^{G}_{t}$ to obtain the context vector $c^{*}$, helping the decoder to decide the position where to copy directly.
\begin{equation}
\setlength\abovedisplayskip{3pt}
\setlength\belowdisplayskip{3pt}
c^{*}=mean(W_{5}h^{U}+W_{6}h^{G}_{t}, dim=-1),
\end{equation}

We obtain the decoder attention $a_{t}$ from the decoder and do some transformation on it. Let it multiply by $P_{copy}$ and get the new attention $\tilde{a}_{t}$
\begin{equation}
\setlength\abovedisplayskip{3pt}
\setlength\belowdisplayskip{3pt}
\tilde{a}_{t}=mean(a_{t},dim=1) \otimes P_{copy},
\end{equation}

Finally, multiply $\tilde{a}_{t}$, $P_{copy}$ and $c^{*}$, and we get the final copy generation $\tilde{P}_{copy}$.
\begin{equation}
\setlength\abovedisplayskip{3pt}
\setlength\belowdisplayskip{3pt}
\tilde{P}_{copy}=c^{*} \otimes \tilde{a}_{t} \otimes P_{copy} 
\end{equation}

\subsubsection{Summary Generating}
To integrate the utterance-level embeddings into the summary generation process, we employ a dual cross-attention mechanism in the model\cite{dualattention2022}.

We first apply the self-attention on the masked output summary embeddings, and then use the output embedding $h^{O}$ to cross-attend to the token-level dialogue hidden states $h^{U}$ produced by the utterance encoder, and get the $P^{q}$:
\begin{equation}
\setlength\abovedisplayskip{3pt}
\setlength\belowdisplayskip{3pt}
P^{q}=MHAtt(h^{O},h^{U}),
\end{equation}
where MHAtt is the standard multi-head attention layer. In a similar way, we apply a multi-head graph attention layer to the utterance-based hidden states $h^{G}_{u}$ and get the $P^{s}$, which aggregate useful knowledge from the updated graph nodes according to the state of each decoding step:
\begin{equation}
\setlength\abovedisplayskip{3pt}
\setlength\belowdisplayskip{3pt}
P^{s}=MHAtt(P^{q},h^{G}_{u}),
\end{equation}

Eventually, we apply a fully connected feed-forward network on $P^{s}$ and get the generation distribution $P_{gen}$. Then, we combine $P_{gen}$ with $\tilde{P}_{copy}$, and obtain the final distribution over the vocabulary of the generated summary.
\begin{equation}
\setlength\abovedisplayskip{3pt}
\setlength\belowdisplayskip{3pt}
P_{prob}=\tilde{P}_{copy} + (1-\tilde{P}_{copy}) \otimes P_{gen}.
\end{equation}

\subsection{Training Objective}
We optimize all parameters of MLEGT by employing the cross-entropy loss between the generated summary and the ground truth summary as the training objective. And we initialize the relevant parameters in our Transformer-based utterance encoder and summary generator with the parameters from the pre-trained language model BART.
\section{Experiments}
\subsection{Experimental Setup}
\subsubsection{Datasets}
The datasets used in this experiment are LCSPIRT\footnote{http://eie.usts.edu.cn/prj/NLPoSUST/LcsPIRT.htm}\cite{LcsPIRT} and Medical DS\footnote{www.sdspeople.fudan.edu.cn/zywei/data/acl2018-mds.zip}\cite{medical}. LCSPIRT is a large-scale Chinese text summarization corpus containing 38,500 text-summary pairs derived from police inquiry dialogue records,
and the texts are deidentified to ensure privacy. Medical DS, on the other hand, consists of 3,260 dialogue-summary pairs collected from the pediatric department of a Chinese online healthcare community. For both datasets, we use the originally proposed train, validation, and test splits, comprising 70\%, 10\%, and 20\% of the data, respectively.

\subsubsection{Implementation Details}
We use BART as the token-level encoder, and the outputs are utilized into the follow utterance-level encoder as well. We apply Adam with a learning rate 1e-5 for BART and the training epoch is set to 15. All Transformer blocks have 4 layers, 8 heads, 1024 hidden units. All hyper-parameters tuning is conducted on the validation set. 

\begin{table}[t]
\caption{Rouge-1, Rouge-2 and Rouge-L scores for different models on LCSPIRT and Medical DS test set. The numbers with * indicate that the improvement of our model over all baselines is statistically significant with $p<0.05$ under t-test.}
\centering
\scalebox{0.78}{
\label{tab1}
\begin{tabular}{l|c|c|c|c|c|c}
\hline
\multirow{2}{*}{\bf Model} & 
\multicolumn{3}{c|}{\bf LCSPIRT}& 
\multicolumn{3}{c}{\bf Medical DS}\\ 
\cline{2-7}& \textbf{Rouge-1} & 
\textbf{Rouge-2} & \textbf{Rouge-L} & \textbf{Rouge-1}  & \textbf{Rouge-2} & \textbf{Rouge-L}\\ 
\hline
PGN & 40.03 & 25.14 & 31.7 & 62.7 & \textbf{55.77} & 57.23 \\
BART & 73.36 & 56.34 & 71.43 & 63.53 & 51.15 & 56.55 \\
T5-pegasus & 77.27 & 69.10 & 72.35 & 58.08 & 51.34 & 52.01 \\
TDS+NTM & 74.78 & 60.37 & 72.84 & 63.85 & 51.82 &  56.53\\
GPT-4o (few-shot) & 80.04 & 67.56 & 78.59 & 61.85 & 45.39 & 55.25 \\
SDDS & 82.27 & 71.99 & 79.13 & \underline{64.9} & 52.19 &  58.42\\
\hline
    MLEGT (Ours) &  \textbf{85.77*} & \textbf{75.47*} & \textbf{83.34*} & \textbf{65.09} & \underline{52.57} & \textbf{58.81*} \\ 
\quad  w/o global topic & 83.96 & 73.25 & 80.69 & 64.82 & 52.08 & \underline{58.43}   \\ 
\quad 
w/o pointer & \underline{85.47} & \underline{74.34} & \underline{82.84} & 64.34 & 52.17 & 57.87   \\ 

\hline

\end{tabular}}
\end{table}

\subsubsection{Baselines}
We compare MLEGT with following baselines: {\bf PGN}\cite{pointer2017}, {\bf BART}\cite{bart2020}, {\bf T5-pegasus}\cite{DBLP:conf/icml/ZhangZSL20}, {\bf TDS+NTM}\cite{zou2021}, {\bf GPT-4o (few-shot)}\cite{gpt-4o} and {\bf SDDS}\cite{sdds2023}. We use the SDDS without the discourse parsing toolkit for fair comparison.

\subsection{Experiments Results}
\subsubsection{Main Results}
We assessed the quality of summarization using Rouge F1 score, including Rouge-1, Rouge-2, and Rouge-L metrics\cite{DBLP:conf/acl/LinO04}. Table I presents the results on the test set for various models. We highlight \textbf{best} and \underline{second best} results. Notably, our model surpasses all compared baselines, achieving state-of-the-art performance across all metrics. Specifically, on the LCSPIRT dataset, our model exhibits significant improvements, up by +3.5\% in ROUGE-1, +3.48\% in ROUGE-2, and +4.21\% in ROUGE-L scores. We also find that our MLEGT can achieve consistently better performance than the strong baselines on Medical DS. To our surprise, PGN that performs poorly on LCSPIRT gets the best ROUGE-2 score on Medical DS. We conjecture the main reason is that there are numerous
medical terms scattered throughout the long conversation structure in the medical dataset, and the pointer mechanism is good at capturing such information while generating
summary.
 
\subsubsection{Ablation Study}
We further investigate the effect of each mechanism of MLEGT by removing the global node and the pointer mechanism, respectively. The results are also shown in Table I. The results of ``w/o global topic'' and ``w/o pointer'' demonstrate a notable decline on both LCSPIRT and Medical DS datasets. This decline underscores the critical role of the proposed global topic as well as the global topic guided pointer mechanism.

\subsubsection{Human Evaluation}

We conducted human evaluation to qualitatively evaluate the generated summaries. As our research focuses on task-oriented dialogue summarization, the informativeness is more critical than the conciseness and fluency for the generated summaries. We focus on how well the generated summaries covered the most important content and compare the accuracy of entity generation across different methods. We randomly selected 200 samples from the LCSPIRT test set and asked five human volunteers to manually annotate entity information such as the \textit{Start} and \textit{End Year/Month/Day/Time}, \textit{Character}, and \textit{Place}. Our volunteers then compared the entities in the generated summaries of MLEGT and SDDS with those in the reference summaries to count the correct rate of entity generation. As illustrated in Fig. 3, MLEGT demonstrates better quality in entity recognition, particularly for \textit{End Month}, \textit{End Date}, \textit{End Time}, \textit{Character}, and \textit{Place}. 
It demonstrates that MLEGT can identify the more important parts across the dialogue, thereby increasing the informativeness compared to the baseline.

\setlength{\textfloatsep}{8pt plus 3pt minus 1pt}
\begin{figure}[t]
\centering
\includegraphics[width=3.6in]{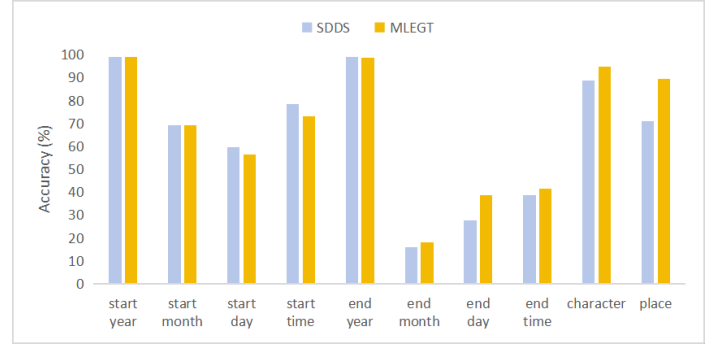}
\caption{The correct rate of entity generation on LCSPIRT.}
\end{figure}

\subsubsection{Case Analysis}
In this subsection, we present an close look for our MLEGT and
the SDDS model via case studies. 
Table II presents a summary example from the LCSPIRT dataset generated by the MLEGT model, in comparison with the baseline SDDS model and the reference. The baseline SDDS model often tends to generate redundant contents and produce some crucial information errors.
Compared to the baseline, our MLEGT model makes fewer errors and captures more critical information throughout the dialogue. 
This further demonstrates the effectiveness of the proposed global topic and the global topic guided pointer mechanism.

\begin{table}[t]
\renewcommand{\arraystretch}{1.2}
\caption{A generated summary case. The underlined part in the reference refers to the entity information, and the {\color{red}red part} in the generated abstract is the wrong generated content.}
\centering
\label{tab:my-table}
\scalebox{0.75}{%
\begin{tabular}{|l|}
\hline

\begin{tabular}[c]{@{}l@{}}
\textbf{Ref Summary}: 
2017年\underline{02月06日08时}到2017年\underline{02月06日17时下午1218时张某某}在 \underline{蝶湖湾} \\ 
\underline{花园1723201} 被嫌疑人以其他\underline{入室手段}窃走\underline{黑色华硕牌A45笔记本电脑1台}。\\ 
From \underline{08:00 on February 6th}, 2017 to \underline{17:00 on February 6th}, 2017 \underline{at 1218 p.m.}, \underline{a black} \\ 
\underline{Asus A45 laptop from Zhang} was stolen by the suspect by other means \underline{in the Butterfly Lake} \\ 
\underline{Bay Garden 1723201}.
\end{tabular} \\ 
\hline

\begin{tabular}[c]{@{}l@{}}
\textbf{SDDS (baseline)}: 2017年02月06日08时到2017年{\color{red}2月06某某}在 {\color{red}昆山市玉山镇蝶湖湾花园 3} \\ 
栋被嫌疑人以其他入室手段窃走1台黑色华硕牌A45型笔记本电脑 {\color{red}......现金人民币500元。} \\ 
{\color{red}某某。银行卡等物品。都是网络通讯软件的。} \\ 
From 08:00 on February 6th, 2017 to {\color{red}February 6th, 2017}, a black Asus A45 laptop from {\color{red}XX} \\ 
was stolen by the suspect by other means in {\color{red}the 3 building of the Butterfly Lake Bay Garden} \\ 
{\color{red}in Yushan Town, Kunshan City}. {\color{red}. . The cash is 500 yuan. Something. Bank cards and other} \\ 
{\color{red}items. It's all from the network communications software.}
\end{tabular} \\ 
\hline

\begin{tabular}[c]{@{}l@{}}
\textbf{MLEGT}: 2017年02月06日08时到2017年{\color{red}03}月06号17时下午1218时张某某在蝶湖湾花园 {\color{red}17232015} \\ 
被嫌疑人以其他入室手段窃走黑色华硕牌A45型笔记本电脑1台。 \\ 
From 08:00 on February 6th, 2017 to 17:00 {\color{red}on March} 6th, 2017, at 1218 PM, a black Asus \\ 
A45 laptop from Zhang was stolen by the suspect by other means in Butterfly Lake Bay \\ 
Garden{\color{red}17232015}.
\end{tabular} \\ 
\hline

\end{tabular}}
\end{table}


\section{Conclusion}
In this paper, we propose a model that utilizes a multi-level encoder with global topic (MLEGT) for task-oriented dialogue summarization,
which enables to discover the temporal and spatial dependencies while obtain a simple yet effective global topic to better capture the crucial information.
Experiments on two public data sets demonstrate the effectiveness of our proposed model in quantitative and qualitative evaluation. 

\section{Acknowledgements}
This work was supported by the National Natural Science Foundation of China (62306119 and 71472068).

\bibliographystyle{IEEEtran}
\bibliography{ref}

\end{CJK}
\end{document}